\documentclass[12pt, titlepage, reqno]{article}

\usepackage{natbib}
\usepackage{apalike}
\usepackage{amssymb, latexsym}
\usepackage{amsmath,amsfonts}
\usepackage{bm}
\usepackage[mathscr]{eucal}
\usepackage{enumerate}
\usepackage{graphicx}
\usepackage{subfigure}
\usepackage{multirow}
\usepackage{color}
\usepackage{amsthm}
\usepackage{url}
\usepackage{tipa}
\usepackage{rotating}
\usepackage{makecell}
\usepackage{titlesec}
\usepackage{lineno}

\titleformat*{\section}{\normalsize\bfseries}

\titleformat*{\subsection}{\normalsize\bfseries}

\newcommand{\vx}{\boldsymbol{x}}
\newcommand{\va}{\boldsymbol{a}}
\newcommand{\sx}{\bar{x}}
\newcommand{\svx}{\bar{\vx}}
\newcommand{\vw}{\boldsymbol{w}}
\newcommand{\vt}{\boldsymbol{t}}
\newcommand{\vth}{\hat{\boldsymbol{t}}}
\newcommand{\tb}{t_b}
\newcommand{\te}{t_e}
\newcommand{\tbh}{\hat{t}_{b}}
\newcommand{\teh}{\hat{t}_{e}}
\newcommand{\taub}{\tau_b}
\newcommand{\taue}{\tau_e}
\newcommand{\vphi}{\boldsymbol{\phi}}
\newcommand{\Xc}{\mathcal{X}}
\newcommand{\Tc}{\mathcal{T}}

\newcommand{\R}{\mathbb{R}}
\newcommand{\E}{\mathbb{E}}

\newcommand{\mytilde}{\raise.17ex\hbox{$\scriptstyle\mathtt{\sim}$}} \renewcommand{\eqref}[1]{Eq.~(\ref{#1})}
\newcommand{\secref}[1]{Section~\ref{#1}}
\newcommand{\figref}[1]{Figure~\ref{#1}}
\newcommand{\tabref}[1]{Table~\ref{#1}}

\newcommand{\indicator}{\bm{1}}

\DeclareMathOperator*{\argmin}{arg\,min}
\DeclareMathOperator*{\argmax}{arg\,max}

\begin{document}

%%% TITLE PAGE
\begin{titlepage}
\begin{center}

\textbf{Automatic measurement of vowel duration via structured prediction}\\

\vspace{5ex}

Yossi Adi, 
Joseph Keshet \\
Department of Computer Science, Bar-Ilan University, Ramat-Gan, Israel,  52900\\
\vspace{5ex}
Emily Cibelli,
Erin Gustafson\\
Department of Linguistics, Northwestern University, Evanston, IL, 60208\\
\vspace{5ex}
Cynthia Clopper\\
Department of Linguistics, Ohio State University, Columbus, OH, 43210\\
\vspace{5ex}
Matthew Goldrick\footnote{e-mail: matt-goldrick@northwestern.edu}\\
Department of Linguistics, Northwestern University, Evanston, IL, 60208\\
\end{center}
\markboth{\hfill Automatic measurement of vowel duration \ }{\hfill Automatic measurement of vowel duration \ }

\end{titlepage}

\begin{abstract}
A key barrier to making phonetic studies scalable and replicable is the need to rely on subjective, manual annotation. To help meet this challenge, a machine learning algorithm was developed for automatic measurement of a widely used phonetic measure: vowel duration. Manually-annotated data were used to train a model that takes as input an arbitrary length segment of the acoustic signal containing a single vowel that is preceded and followed by consonants and outputs the duration of the vowel. The model is based on the structured prediction framework. The input signal and a hypothesized set of a vowel's onset and offset are mapped to an abstract vector space by a set of acoustic feature functions. The learning algorithm is trained in this space to minimize the difference in expectations between predicted and manually-measured vowel durations. The trained model can then automatically estimate vowel durations without phonetic or orthographic transcription. Results comparing the model to three sets of manually annotated data suggest it out-performed the current gold standard for duration measurement, an HMM-based forced aligner (which requires orthographic or phonetic transcription as an input).
\end{abstract}

%%%%%%%%%%%%%%%%%%%%%%%%%%%%%%%%%%%%%%%%%%%%%%%%%%%%%%%%%%%%%%%%%%%%%%%%%%%%%%
%
% DON'T DELETE - these are the correct PACS code for our submission:
%
% PACS, the Physics and Astronomy Classification Scheme.
%\pacs{43.70.Jt, 43.72.Ar, 43.72.Lc, 43.70.Fq}
% these are: 
% - Instrumentation and methodology for speech production
% research
% - Speech analysis and analysis techniques; parametric representation of speech
% - Time and frequency alignment procedures for speech
% - Acoustical correlates of phonetic segments and suprasegmental properties: stress, timing, and intonation
%
%%%%%%%%%%%%%%%%%%%%%%%%%%%%%%%%%%%%%%%%%%%%%%%%%%%%%%%%%%%%%%%%%%%%%%%%%%%%%%

\addtocounter{page}{2}

%%%%%%%%%%%%%%%%%%%%%%%%%%%%%%%%%%%%%%%%%%%%%%%%%%%%%%%%%%%%%%%%%%%%%%%%%%%%%%

\section{INTRODUCTION}
\label{sec:introduction}

\setlength{\parindent}{5ex}

Understanding the factors that modulate vowel duration has been a long-standing focus of laboratory research in acoustic phonetics \citep{peterson1960duration}. The vast majority of such research--from the mid-20th century to the start of the 21st--has relied on manual annotation to determine vowel duration. There are two key issues with such an approach. Given the labor intensive nature of such annotation, there are substantial limitations on the amount of data that can be practically analyzed; this limits the statistical power and types of issues that phonetic studies can address. Furthermore, given the fundamental reliance on subjective annotator judgments, analyses cannot be directly replicated by other researchers.  

An alternative to this approach is to pursue algorithms for automatic alignment of vowel boundaries. The current standard approach for vowels is to utilize forced alignment algorithms \citep{yuan2013automatic}. However, this approach suffers from two important limitations: it requires an orthographic transcription, and frequently requires substantial preprocessing of the data to insure adequate performance \citep{evanini2009permeability}. 

In this paper we propose a method for automatic measurement of the duration of single vowels using structured prediction techniques which have provided excellent results in analyzing other phonetic measures \citep{KeshetShSiCh07, SondereggerKe12}. The algorithm was trained at the segment level on manually annotated data to extract the vowel start and end times; this provides a straightforward way to compute the duration of the vowel. Following the structure of the vast majority of laboratory studies of vowel duration, we assume the input signal contains a single vowel, proceeded and followed by a consonant (CVC)--no additional detailed information about the phonetic transcription is required to process the speech. This tool will allow laboratory studies to rapidly and reliably examine how the duration of vowels in monosyllabic words varies across participants and elicitation contexts.

We evaluated our method on data from three phonetic studies: one focusing on vowel duration \citep{heller2014grammatical}, the second using vowel segmentation to automatically determine points for formant analysis \citep{clopper2014effects}, and the third a standard set of vowel production norms  \citep{hillenbrand1995acoustic}.  We compared results with our model to the state-of-the-art in vowel duration measurement, HMM-based forced alignment \citep{rosenfelder2014fave}. We also assessed whether inferential statistical models fit to data from our model and HMM-based forced alignment replicated the patterns obtained from manual data. The results suggest that our algorithm is superior to the current gold standard at matching the manual measurements of vowel duration, both in terms of deviation and in replicating inferential statistical results.

The paper is organized as follows. In \secref{sec:problem_setting} we state the problem definition formally. We then present the learning framework (\secref{sec:learning_framework}), algorithm (\secref{sec:learning_algorithm}), and the acoustic features and feature functions (\secref{sec:features}). In \secref{sec:datasets} we describe the datasets we use to train the models and evaluate the performance of the algorithm. In \secref{sec:methods} we detail the particular methods used to implement our algorithm here, along with the standard approach to vowel duration measurement. Experimental results are detailed in two sections: the first focusing on measurement deviation (\secref{sec:deviation}) and the second on the reproduction of results from inferential statistical models (\secref{sec:evaluation_regression}). We conclude the paper with possible applications and extensions in \secref{sec:discussion}.

%%%%%%%%%%%%%%%%%%%%%%%%%%%%%%%%%%%%%%%%%%%%%%%%%%%%%%%%%%%%%%%%%%%%%%%%%%%%%%
\section{PROBLEM SETTING}
\label{sec:problem_setting}

In the context of a typical laboratory study of speech, the goal of automatic vowel duration measurement is to accurately predict the time difference between the vowel onset and offset, given a segment of the acoustic signal in which the vowel is preceded by a consonant and followed by a consonant. The acoustic sample can be of any length, but should include only one vowel. We assume there is a small portion of silence before and after the uttered speech, but do not require the beginning of the speech signal or the vowel onset to be synchronized with the onset or offset of the acoustic sample.

We turn to describing the problem formally. Throughout the paper we write scalars using lower case Latin letters, e.g.,  $x$, and vectors using bold face letters, e.g., $\vx$. A sequence of elements is denoted with a bar $\sx$ and its length is written as $|\sx|$. Similarly a sequence of vectors is denoted as $\svx$ and its length by $|\svx|$.

\begin{figure}[h]
\centering
\includegraphics[width=0.78\linewidth]{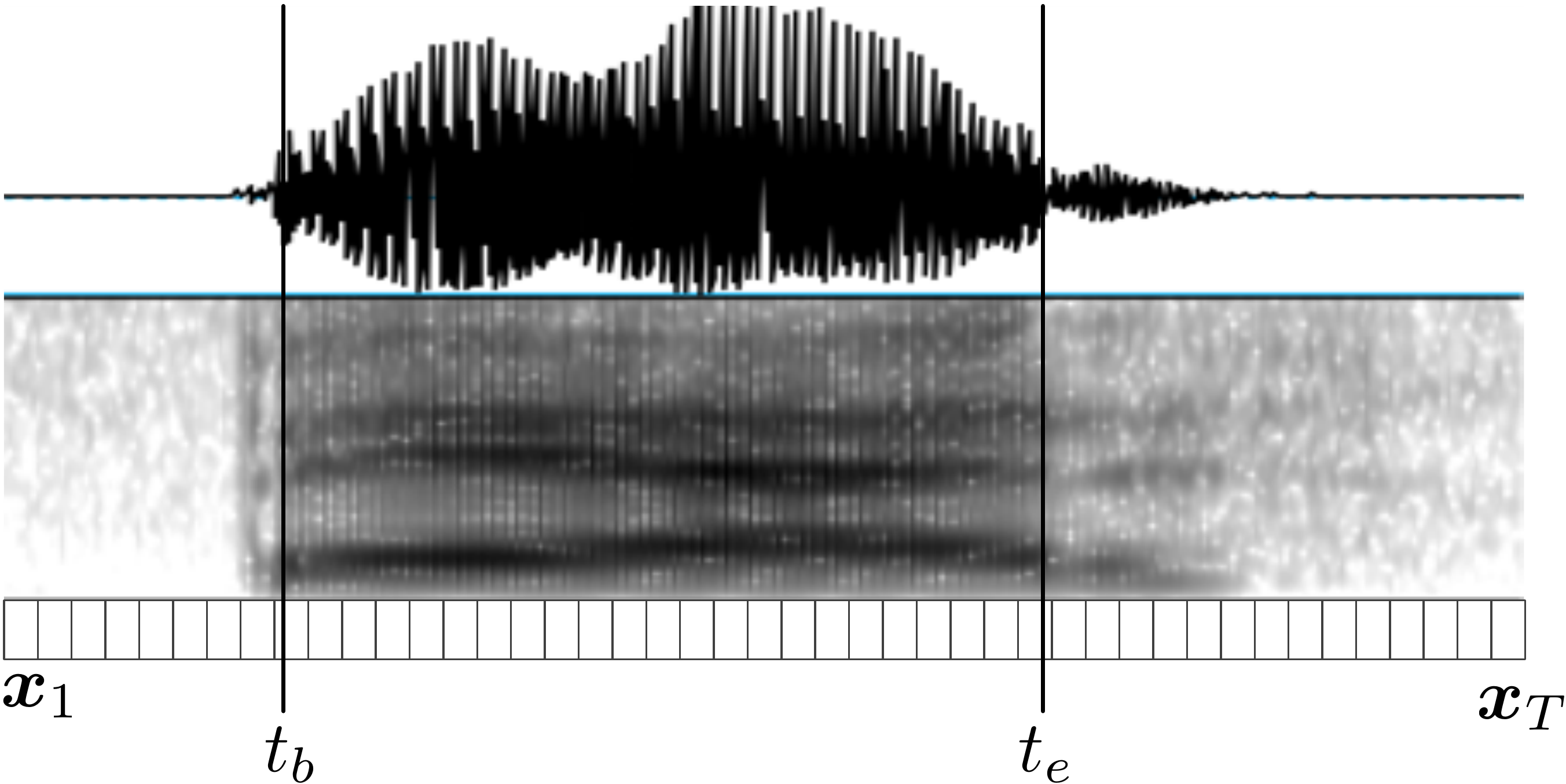}
\caption{An example of our notation. The top panel presents the signal in the time domain, and the middle panel presents the spectrogram of the signal. The vertical solid lines present the annotated vowel's onset $\tb$ and offset $\te$. The speech signal is represented by a sequence of acoustic feature vectors as depicted in the lower panel. For the $t$-th frame, the acoustic feature vector is denoted by $\vx_t$.}
\label{fig:vowel_notation}
\end{figure}

The acoustic sample is represented by a sequence of acoustic feature vectors denoted by $\svx = (\vx_1, \vx_2,...,\vx_T)$, where each $\vx_t$ ($1 \leq t \leq T$) is a $d$-dimensional vector that represents the acoustic content of the $t$-th frame. The domain of the feature vectors is denoted as $\Xc \subset \R^d$. The input acoustic sample can be of an arbitrary length, thus $T$ is not fixed. We denote by $\Xc^*$ the set of all finite length segments of acoustic signal over $\Xc$. In addition, we denote by $\tb\in\Tc$ and $\te\in\Tc$ the vowel onset and offset times in frame units, respectively, where $\Tc = \{1,...,T\}$, and the total duration of the input acoustic sample is $T$ frames. For brevity we denote this pair by $\boldsymbol{t}=(\tb, \te)$, and refer to it as an \emph{onset-offset pair}. Practically there are constraints on $\tb$ and $\te$ and they cannot take any value in $\Tc$, e.g., the vowel onset $\tb$ cannot be $T$ or $T-1$. Our notation is depicted in \figref{fig:vowel_notation}.

%%%%%%%%%%%%%%%%%%%%%%%%%%%%%%%%%%%%%%%%%%%%%%%%%%%%%%%%%%%%%%%%%%%%%%%%%%%%%% Figure 1

%%%%%%%%%%%%%%%%%%%%%%%%%%%%%%%%%%%%%%%%%%%%%%%%%%%%%%%%%%%%%%%%%%%%%%%%%%%%%%

Our goal is to find a function $f$ from the domain of segments of acoustic signal, $\Xc^*$, to the domain of all onset-offset pairs, $\Tc^2$. Given a segment of the acoustic signal $\svx$, let $\vth = f(\svx)$ be the predicted onset-offset pair, where $\vth=(\tbh, \teh)$. The quality of the prediction is assessed using a \emph{loss function}, denote by $\gamma(\vt,\vth)$, that measures the magnitude of the penalty when predicting the pair $\vth$ rather than the target pair $\vt$. Formally, $\gamma: \Tc^2\times\Tc^2 \rightarrow \R_+$ is a function that receives as input two ordered pairs, and returns a positive scalar. We assume that if both the predicted and the target pairs are the same, then the loss is zero, $\gamma(\vt,\vt) = 0$. 

%%%%%%%%%%%%%%%%%%%%%%%%%%%%%%%%%%%%%%%%%%%%%%%%%%%%%%%%%%%%%%%%%%%%%%%%%%%%%%
\section{LEARNING FRAMEWORK}
\label{sec:learning_framework}

We assume that an input acoustic sample and a target onset-offset pair are drawn from a fixed but unknown distribution $\rho$ over the domain of the segments of acoustic signal and the vowel onset-offset pairs, $\mathcal{X}^* \times \Tc^2$. We define the \emph{risk} of $f$ as the expected loss when using $f$ to predict the onset-offset pair of the acoustic sample $\svx$, that is,
\begin{equation}
\label{eq:risk}
R(f) = \E_{(\svx, \vt)\sim \rho}[\gamma (\vt,f(\svx))],
\end{equation} 
where the expectation is taken with respect to the input acoustic sample $\svx$ and the annotated onset-offset pair $\vt$ drawn from $\rho$. Our goal is to find $f$ that minimizes the risk. Unfortunately, this cannot be done directly since $\rho$ is unknown. Instead, we use a training set of examples that served as a restricted window through which we can estimate the quality of the prediction function according to the distribution of unseen examples in the real world. The examples are assumed to be identically and independently distributed (i.i.d.) according to the distribution $\rho$. 

Each example in the training set is composed of a segment of the acoustic signal and a manually annotated vowel onset and offset pair. The manual annotations are not exact, and naturally depend both on human errors as well as objective difficulties in placing the vowel boundaries, e.g., between the vowel and a sonorant. Hence in this work we focus on a loss function that inherently takes into account the discrepancy in the annotations. That is,
\begin{equation}
\label{eq:loss}
\gamma (\vt,\vth) =  \big[|\tbh -  t_b| - \taub\big]_+ + \big[|\teh -  t_e| - \taue\big]_+,
\end{equation} 
where $[\pi]_+=\max\{0,\pi\}$, and $\taub$, $\taue$ are pre-defined parameters. The above function measures the absolute differences between the predicted and the target vowel onsets and offsets. It allows a mistake of $\taub$ and $\taue$ frames at the onset and offset of the vowel respectively, and only penalizes predictions that are greater than $\taub$ or $\taue$ frames.

Our learning model belongs to the structured prediction framework. In this framework it is assumed that the output prediction is complex and has some internal structure. In our case, the vowel onset and vowel offset times are related and dependent, e.g., the vowel has a typical duration that depends on the vowel onset and offset.

In the structured prediction model, the function $f$ is based on a fixed mapping $\vphi: \Xc^* \times \Tc^2 \rightarrow \R^n$ from the set of segments of acoustic signal and target onset-offset pairs to a real vector of length $n$; we call the elements of this mapping \emph{feature functions} or \emph{feature maps}. Intuitively, the feature functions represent our knowledge regarding good locations of the onset or the offset of the vowel within the acoustic signal. For example, consider \figref{fig:vowel_notation}. It can be seen that the spectral change is high at the areas of $\tb$ and $\te$. Based on this, we can adopt a feature function based on the distance between the spectrum a frame before and a frame after the presumed $\tb$. This function is going to be high if the presumed $\tb$ is in the vicinity of the actual target vowel onset and is going to be low at a random place in the acoustic signal.

Our prediction function is a linear decoder with a vector of parameters $\vw \in \R^n$ that is defined as follows:
\begin{equation}
\label{eq:decoding}
f_{\vw}(\svx) = \argmax_{\vth \in \Tc^2} ~ \vw^{\top} \vphi(\svx, \vth).
\end{equation} 
The subscript $\vw$ is added to the function $f$ to stress that it depends on the weight vector $\vw$.

Ideally, we would like our learning algorithm to find $\vw$ such that the prediction minimizes the loss on unseen data. Recall, we assume there exists some unknown probability distribution $\rho$ over pairs $(\svx,\vt)$. We would like to set $\vw$ so as to minimize the expected loss, or the \emph{risk}, for predicting $f_{\vw}(\svx)$,
\begin{equation}
\label{eq:w*}
\vw^* = \argmin_{\vw} ~ \E_{(\svx,\vt) \sim \rho} [\gamma(\vt, f_{\vw}(\svx))]. 
\end{equation}
It is hard to directly minimize this objective function since $\rho$ is unknown and the loss $\gamma$ is often a combinatorial non-convex function \citep{Keshet14}. In the next section we describe the learning algorithm that aims at minimizing the risk, and then we describe the set of feature functions in \secref{sec:features}.

%%%%%%%%%%%%%%%%%%%%%%%%%%%%%%%%%%%%%%%%%%%%%%%%%%%%%%%%%%%%%%%%%%%%%%%%%%%%%%
\section{DIRECT LOSS MINIMIZATION (DLM) ALGORITHM}
\label{sec:learning_algorithm}

Recall that our goal is to directly optimize the objective in \eqref{eq:w*}. Unfortunately, if the output space is discrete we can not use direct gradient decent since the loss $\gamma(\vt,f_{\vw}(\svx))$ is not a differentiable function of $\vw$ \citep{Keshet14}.  \cite{McAllesterHaKe10} showed that if the input space $\Xc^*$ is continuous, we can compute the gradient of the expected loss, i.e., the risk, in \eqref{eq:w*} even when the output space is discrete in terms of the feature functions. Specifically the gradient can be expressed in a closed form solution as follows: 
\begin{equation}
\label{eq:loss_thm}
\nabla_{\vw} \E\Big[\gamma(\vt, f_{\vw}(\svx) \Big] = 
\lim_{\epsilon \rightarrow 0} \frac{\E \Big[\vphi(\svx,f^{\epsilon}_{\vw}(\svx)) - \vphi(\svx,f_{\vw}(\svx))\Big]}{\epsilon},
\end{equation}
where the expectation on both sides is with respect to the tuple $(\svx,\vt)$ drawn from $\rho$, and $f^{\epsilon}_{\vw}$ is defined as follows:	
\begin{equation}
f^{\epsilon}_{\vw}(\svx) = \argmax_{\vth\in\Tc^2} ~\vw^\top\vphi(\svx,\vth) + \epsilon \,\gamma(\vt,\vth).
\end{equation}
Using stochastic gradient decent we get the following update rule:
\begin{equation}
\label{eq:update_rule}
\vw_{t+1} = \vw_{t} + \frac{\eta_t}{\epsilon}\left(
\vphi(\svx,f_{\vw}(\svx))  - \vphi(\svx,f^{\epsilon}_{\vw}(\svx))
\right),
\end{equation}
where $\eta_t$ is the learning rate. At training time, we set $\eta_t=\eta_0/\sqrt{t}$, where $\eta_0$ is a parameter and $t$ is the iteration number, and we set $\epsilon$ as a fixed small parameter, which is selected from a held-out development set. The actual values of the parameters are detailed in \secref{sec:deviation}.

\begin{figure}[h]
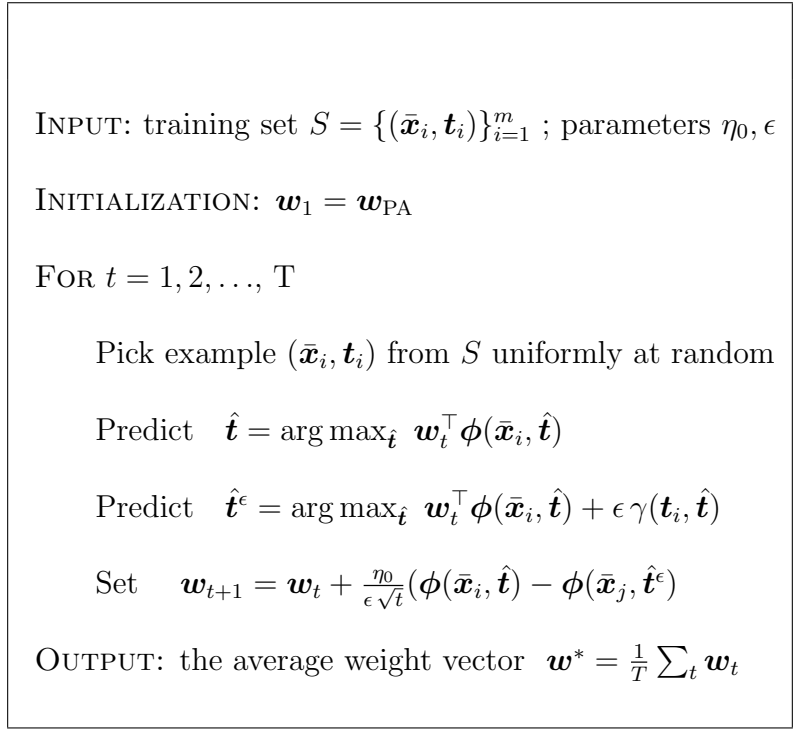

\centerline{
\begin{tabular}{|c|}
\hline \\
\parbox[t]{\linewidth}{
\begin{tabbing}
~\=\textsc{Input:}~\=training set $S=\{(\svx_{i}, \vt_{i})\}_{i=1}^{m}$ ; parameters $\eta_0, \epsilon$ \\ 
\> \textsc{Initialization:} $\vw_1=\vw_{\text{PA}}$ \\ 
\> \textsc{For}\=~\= $t=1,2,\ldots$, T \\ 
\>\> Pick example $(\svx_{i}, \vt_{i})$ from $S$ uniformly at random \\ 
\>\> Predict ~~$\vth = \argmax_{\vth} ~\vw_{t}^\top\vphi(\svx_i,\vth)$ \\ 
\>\> Predict ~~$\vth^{\epsilon} = \argmax_{\vth} ~\vw_{t}^\top\vphi(\svx_i,\vth) + \epsilon \,\gamma(\vt_{i},\vth)$ \\ 
\>\> {Set} ~~ $\vw_{t+1} = \vw_{t} + \frac{\eta_0}{\epsilon\,\sqrt{t}}(\vphi(\svx_{i}, \vth) - \vphi(\svx_{j}, \vth^{\epsilon})$ \\
~\textsc{Output:} the average weight vector $~ \vw^* = \frac{1}{T}\sum_t \vw_t$
\end{tabbing} 
} \\
\hline
\end{tabular}
}
\caption{Direct loss minimization training procedure.}  
\label{fig:dlm_algorithm}
\end{figure}

Since the objective in \eqref{eq:loss_thm} is not a convex function in the model parameters $\vw$,  gradient descent is not guaranteed to find the optimal parameter settings; it may converge to a local minimum. We initialize the model parameters with a weight vector of parameters that was pre-trained using the structured prediction passive-aggressive (PA) algorithm \citep{CrammerDeKeShSi06}. The vector of parameters that was obtained from the PA training is denoted by $\vw_{\text{PA}}$. A pseudo code of the training algorithm is given in \figref{fig:dlm_algorithm}.

%%%%%%%%%%%%%%%%%%%%%%%%%%%%%%%%%%%%%%%%%%%%%%%%%%%%%%%%%%%%%%%%%%%%%%%%%%%%%%Figure 2

%%%%%%%%%%%%%%%%%%%%%%%%%%%%%%%%%%%%%%%%%%%%%%%%%%%%%%%%%%%%%%%%%%%%%%%%%%%%%%

%%%%%%%%%%%%%%%%%%%%%%%%%%%%%%%%%%%%%%%%%%%%%%%%%%%%%%%%%%%%%%%%%%%%%%%%%%%%%%
\section{FEATURES AND FEATURE FUNCTIONS}
\label{sec:features}

In this section we describe the acoustic features $\vx\in\Xc$ and the feature functions $\vphi(\svx,\vt)$, which were designed specifically for the problem of vowel duration measurement. These features are primarily motivated by the desire to reflect the workflow of the annotators \citep{peterson1960duration}. For example, in \cite{peterson1960duration} the beginning of the final voiced plosives was determined by comparing narrow-band and broad-band spectrograms and look for the moment in time when the energy in the higher harmonics is greatly diminished. In a similar fashion, we have acoustic features of narrow-band and broad-band energies, and a set of feature functions that allow the machine learning algorithm to compare and weight them, respectively. Moreover, after an empirical evaluation of the performance, and analyzing the source of various errors, we added features which related to the predictions of the phoneme classifiers (e.g., vowel, nasal, etc.).

\subsection{Acoustic Features}

The main function of the acoustic features representation is to preserve the crucial information-bearing elements of the speech signal and to suppress irrelevant details. We extracted $d=16$ acoustic features every 5 ms, in a similar way to the feature set used in \cite{SondereggerKe12}, but with different time spans. The first 5 features are based on the short-time Fourier transform (STFT) taken with a 25 ms Hamming window. The features are short-term energy, $E_{\text{short-term}}$; the log of the total spectral energy, $E_{\text{total}}$; the log of the energy between 20 and 300 Hz, $E_{\text{low}}$; the log of the energy above 3000 Hz, $E_{\text{high}}$; and the Wiener entropy, $H_{\text{wiener}}$, a measure of spectral flatness \citep{SondereggerKe12}. The sixth feature, $S_{\max}$, is the maximum of the power spectrum calculated in a region from 6 ms before to 18 ms after the frame center. The seventh feature, $\hat{F}_{0}$, is the normalized fundamental frequency estimator, extracted using the algorithm of \cite{sha2004multiband} every 5 ms, and smoothed with a Hamming window. The eighth feature is the binary output of a voicing detector based on the RAPT pitch tracker \citep{talkin1995robust}, smoothed with a Hamming window - denoted as $V_{\text{RAPT}}$. The ninth feature is the number of zero crossings in a 5 ms window --- denoted as $N_{\text{ZC}}$.

The next set of acoustic features are based on a pre-trained phoneme classifier's predictions and scores as in \cite{DekelKeSi04a}. This phoneme classifier was trained on the TIMIT dataset \citep{garofolo1993timit}, using the standard MFCC features with a passive aggressive classifier \citep{CrammerDeKeShSi06}. The tenth acoustic feature is an estimated probability of whether a vowel is uttered at the input frame. The feature is a smoothed version of an indicator function that states if the phoneme predicted by the classifier at the current frame is a vowel: $\indicator[\hat{y}_t\in\textbf{vowels}]$, where $\hat{y}_t$ is the phoneme predicted by the phoneme classifier at time $t$ and $\textbf{vowels}$ is the set of all vowels. The eleventh feature is defined to be the same as the tenth feature, but is for nasal phonemes. These features are denoted as $G_{\text{vowel}}$ and $G_{\text{nasal}}$, respectively. The twelfth feature is the likelihood of a vowel at the current time frame, $L_{\text{vowel}}$. The likelihood is computed as the Gibbs measure of the phoneme classifier's scores for all the vowel phonemes normalized by the total score for all phonemes.

The last four features are based on the spectral changes between adjacent frames, using Mel-frequency cepstral coefficients (MFCCs) to represent the spectral properties of the frames. Define by $D_j=d(\va_{t-j},\va_{t+j})$ the Euclidean distance between the MFCC feature vectors $\va_{t-j}$ and $\va_{t+j}$, where $\va_t\in\R^{39}$ for $1 \le t \le T$. The features are denoted by $D_j$, for $j\in \{1, 2, 3, 4\}$.

\begin{figure}[h!]
\centering
\includegraphics[width=0.7\linewidth]{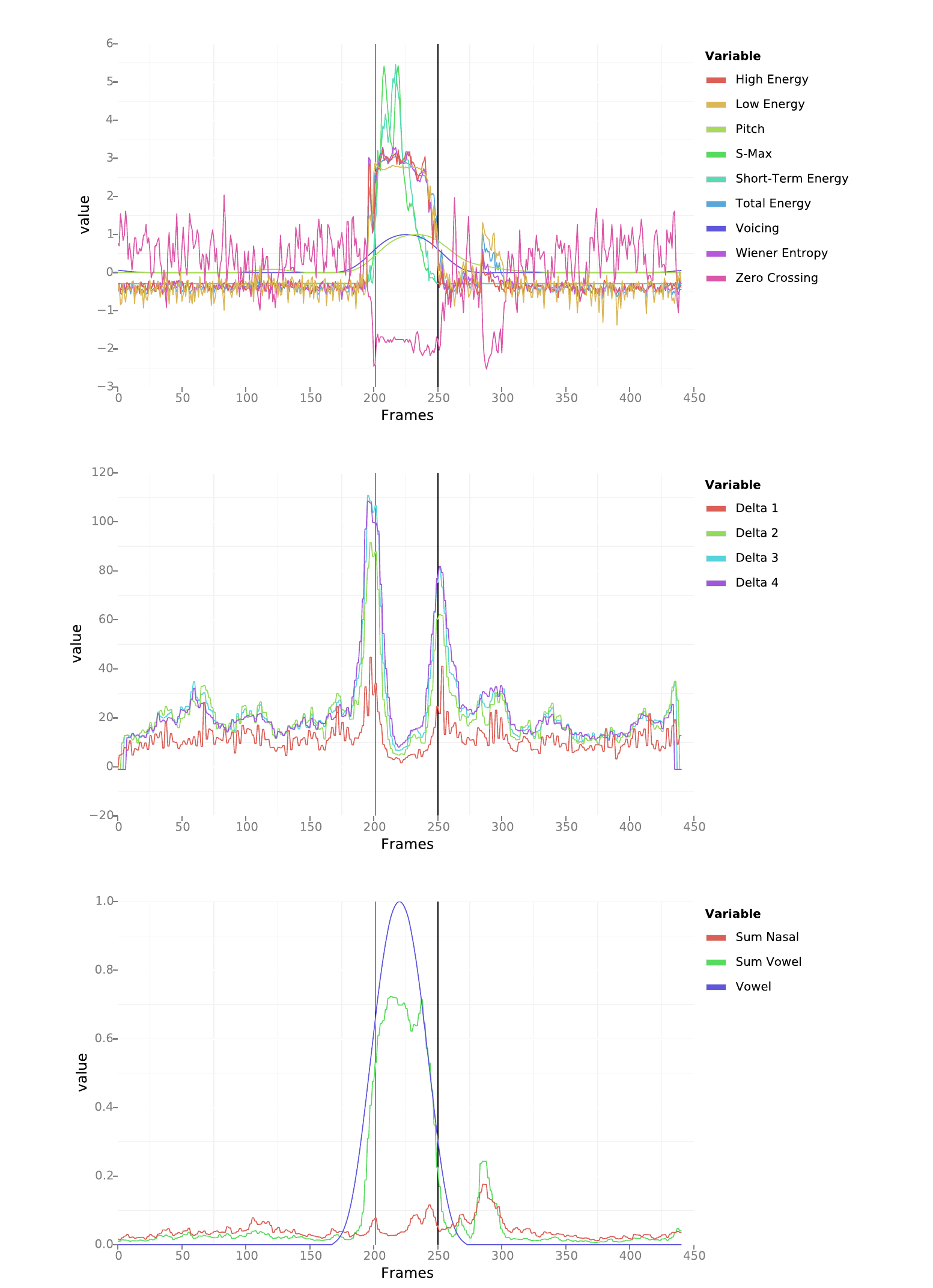}
\caption{Values of acoustic features for an example acoustic sample (the word ``got''). The vertical dashed lines indicate the annotated onset and offset of the vowel (color online).}
\label{fig:acoustic_features}
\end{figure}

\figref{fig:acoustic_features} shows the trajectories of the features for a typical vowel in a CVC context (the word ``got'').

%%%%%%%%%%%%%%%%%%%%%%%%%%%%%%%%%%%%%%%%%%%%%%%%%%%%%%%%%%%%%%%%%%%%%%%%%%%%%%Figure 3
%%%%%%%%%%%%%%%%%%%%%%%%%%%%%%%%%%%%%%%%%%%%%%%%%%%%%%%%%%%%%%%%%%%%%%%%%%%%%%

\subsection{Feature Functions}

We turn now to describe the feature functions. Recall that the feature functions are designed to be correlated with a good positioning of the onset-offset pair, $\vt$, in the acoustic signal, $\svx$. While generally each feature function $\phi_i(\svx,\vt)$, for $1 \le i \le n$, gets as input a sequence of acoustic features, $\svx$, and a presumed onset-offset pair $\vt$, they can practically use only a subset of the acoustic features (e.g., only a sequence over the sixth feature, as opposed to a sequence over any features). Some of the feature functions are based on the average of an acoustic feature $x$ from frame $t_1$ to frame $t_2$ defined as
\begin{equation}
\mu(\sx, t_1, t_2) = \frac{1}{t_2-t_1+1}\sum_{t=t_1}^{t_2} x_t.
\end{equation}
The functions that we identified as being correlated with a good positioning of the onset-offset pair can be divided into four groups based on the structure of the functions; each group was implemented over a set of the above acoustic features (n.b.: a given acoustic feature might be associated with multiple feature functions):

\medskip\noindent\textbf{Type 1:} This type of feature function gets as input a sequence of one of the features, $\sx$, and a time $t$. The time $t$ can represent the onset or the offset of the vowel. Formally, the features of this type are of the form
\begin{equation}
\phi(\sx, t) = x_t
\end{equation}
Feature functions from this type are computed for the acoustic features $E_{\text{total}}$, $E_{\text{low}}$, $E_{\text{high}}$, and $S_{\max}$ at the presumed vowel onset time $\tb$. It is also computed for the acoustic feature $D_j$, $j=1,2,3,4$ for both vowel onset $\tb$ and offset times $\te$. It can be seen from \figref{fig:acoustic_features} that these acoustic features have a high value exactly at $\tb$ or $\te$ or both, respectively.

\medskip\noindent\textbf{Type 2:} The second set of feature functions is a coarse estimation of the derivative around a time frame of interest. Given a sequence of acoustic feature values $\sx$ and a specific time frame $t$, the feature functions of this type compute the difference in the mean of $\Delta$ frames before $t$ and the mean of $\Delta$ frames after $t$. That is 
\begin{equation}
\phi(\sx, t, \Delta) = \mu(\sx, t-\Delta, t-1)
~-~
\mu(\sx, t, t+\Delta-1).
\end{equation}
\tabref{tab:features_type2} describes all the feature functions of this type, for the values of the number of frames processed, $\Delta$, and the acoustic features. For some acoustic features we wanted to take into account the possibility that they might not be coincident with vowel boundary, but occur at an adjacent point in the acoustic signal. We did that by considering the point of interest to be an offset version of the presumed onset or offset. For example, $\tb - 2$ means the feature computes averages before and after the time frame $\tb$ with an offset of 2 frames. It can be seen from \figref{fig:acoustic_features} that indeed the acoustic features described in \tabref{tab:features_type2} have abrupt changes at the specified time frame $t$.

%%%%%%%%%%%%%%%%%%%%%%%%%%%%%%%%%%%%%%%%%%%%%%%%%%%%%%%%%%%%%%%%%%%%%%%%%%%%%%
\begin{sidewaystable}\scriptsize
\caption{The feature functions of Type 2, $\phi(\sx, t_1, \Delta)$. Values in the tables are the time frame $t_1$ used for each acoustic feature (columns) and for each window length (rows). Window lengths 8 and 10 are repeated to indicate cases where the point of interest is offset (see text for details).}
\label{tab:features_type2}
\centering
\centerline{
%\begin{tabular}{ c c c c c c c c c c c c c c c c c} 
\begin{tabular}{ l l l l l l l l l l l l l l l l l} 
\hline
\hline 
$\Delta$ ~& $E_{\text{short-term}}$ ~& $E_{\text{low}}$ ~& $E_{\text{high}}$ ~& $E_{\text{total}}$ ~& $H_{\text{wiener}}$ ~& $S_{\max}$ ~& $\hat{F}_{0}$ ~& $V_{\text{RAPT}}$ ~& $N_{\text{ZC}}$ ~& $G_{\text{vowel}}$ ~& $G_{\text{nasal}}$ ~& $L_{\text{vowel}}$ ~& $D_1$ ~& $D_2$ ~& $D_3$ ~& $D_4$ \\
\hline
 1 & & & & & & $\tb$ & & & & & & & & & &\\
2 & & & & & & $\tb$ & & & & & & & & & &\\
3 & $\tb, \te$ & & & & & $\tb$& & & & & & &$\tb, \te$&$\tb, \te$&$\tb, \te$&$\tb, \te$\\
4 & $\tb, \te$ & & & & & $\tb$& & & & & & & & & &\\
5 & $\tb$ & & & & & $\tb$& & & & & & & & & &\\
6 & & $\tb$ & & & & & & & &$\tb, \te$& &$\tb, \te$ & & & &\\
8 & &$\tb, \te$ & $\tb, \te$& $\tb, \te$&$\tb, \te$& &$\tb, \te$&$\tb, \te$&$\tb, \te$& $\tb, \te$& $\te$ &$\tb, \te$ & & & &\\
8 & &$\tb-2$ & $\tb-2$& $\tb-2$ & & & & & & & & & & & &\\
10 & &$\tb, \te$&$\tb, \te$&$\tb, \te$&$\tb, \te$& &$\tb, \te$&$\tb, \te$&$\tb, \te$& $\tb, \te$& $\te$ &$\tb, \te$ &$\tb, \te$&$\tb, \te$&$\tb, \te$&$\tb, \te$\\
10 & &$\tb-4$ & $\tb-4$& $\tb-4$ & & & & & & & & & & & &\\
\hline               
\hline 
\end{tabular}
}
\end{sidewaystable}
%%%%%%%%%%%%%%%%%%%%%%%%%%%%%%%%%%%%%%%%%%%%%%%%%%%%%%%%%%%%%%%%%%%%%%%%%%%%%%

\medskip\noindent\textbf{Type 3:} The third type of feature function is an extension of the second type. Rather than considering the average of an acoustic feature before and after a single time frame, we refer to the average between the presumed onset and offset times. Feature functions of this type return two values: (i) the average between the presumed onset and offset times minus the average of $\Delta$ frames before $\tb$; and (ii) the average between the presumed onset and offset times minus the average of $\Delta$ frames after the offset time $\te$. Those two values correspond to two elements in the vector 
\begin{equation}\displaystyle
\vphi(\sx, \tb, \te, \Delta) =
\left[ \begin{array}{c}
\mu(\vx,\tb,\te)
-
\mu(\vx,\tb-\Delta,\tb-1) \\
\mu(\vx,\tb,\te)
-
\mu(\vx,\te+1,\te+\Delta)
\end{array} \right].
\end{equation}
This type of feature function is computed for the acoustic features $E_{\text{short-term}}$, $E_{\text{low}}$, $E_{\text{high}}$, $E_{\text{total}}$, $V_{\text{RAPT}}$, $N_{\text{ZC}}$, and $L_{\text{vowel}}$. Again, it can be see from \figref{fig:acoustic_features} that these features have high (or low) mean values between the time frames of interest. 

\medskip\noindent\textbf{Type 4:} The fourth type of feature function is oblivious to the acoustic signal and returns a probability score for the presumed vowel duration, $\te-\tb$. We have two feature functions of this type. The first one is based on the Normal distribution with parameters $\hat{\mu}$, $\hat{\sigma}^2$ that are estimated from the training data:
\begin{equation}
\phi(\tb,\te) = \mathcal{N}(\te-\tb; \hat{\mu}, \hat{\sigma}^2).
\end{equation}
Similarly, the second feature function of this form is based on the Gamma distribution with parameters $\hat{k}$ and $\hat{\theta}$, that are estimated from the training set:
\begin{equation}
\phi(\tb,\te) = \Gamma(\te-\tb; \hat{k}, \hat{\theta}).
\end{equation}
While the Gamma distribution is an appropriate distribution to describe the vowel duration alone, we found empirically that adding the Normal distribution improves performance. 

%%%%%%%%%%%%%%%%%%%%%%%%%%%%%%%%%%%%%%%%%%%%%%%%%%%%%%%%%%%%%%%%%%%%%%%%%%%%%%

\section{DATASETS}\label{sec:datasets}

In order to get reliable results we used three different datasets to evaluate the performance of our system. In this section we will give a short description of each dataset. 

\subsection{Heller and Goldrick, 2014 (HG)}

This corpus \citep{heller2014grammatical} is drawn from a study investigating how grammatical class constraints influence the activation of phonological neighbors. The influence of neighbors on phonetic processing was indexed by vowel durations. It contains segments of acoustic signal from 64 native English speakers (55 female) aged 18-34 with no history of speech or language deficits. Participants were first familiarized to a set of pictures along with their intended labels. They were then asked to name aloud the noun depicted by a picture in two contexts: when the picture was presented alone, and when the picture occurred at the end of a non-predictive sentence. Participants were instructed to produce the name as quickly and accurately as possible. Trials with errors or disfluencies were excluded, along with two items with high error rates. In addition, data from three subjects reported in the original paper were excluded from the present analysis, as they did not consent to public use of their data. The remaining 2395 recorded segments of acoustic signal contain one English CVC noun with vowels /i, \textipa{E}, \textipa{ae}, \textipa{A}, o\textipa{U}, u/. 

\subsection{Clopper and Tamati, 2014 (CT)}

The second dataset \citep{clopper2014effects} contains segments of acoustic signal from 20 female native English speakers aged 18-22 with no history of speech or language deficits. The participants were evenly split between two American English dialects (Northern and Midland). As part of a larger study \citep{clopper2002indiana}, participants read aloud a list of 991 CVC words. This study focused on 39 target words (777 tokens) which did or did not have a lexical contrast between either /\textipa{E}/ vs. /\textipa{ae}/ (e.g., dead-dad vs. deaf-*daff) or /\textipa{A}/ vs. /\textipa{O}/ (e.g., cot-caught vs. dock-*dawk). Words with a lexical contrast are referred to as \textit{competitor} items and those without are referred to as \textit{no competitor} items. 

\subsection{Hillenbrand, Getty, Clark, and Wheeler, 1995 (HGCW)}

The third dataset consists of data from a laboratory study conducted by \cite{hillenbrand1995acoustic}. It contains segments of acoustic signal from 45 men, 48 women, and 46 ten-to 12-year-old children (27 boys and 19 girls). 87\% of the participants were raised in Michigan, primarily in the southeastern and southwestern parts of the state. 
	The audio recordings contain 12 different vowels (/\textipa{i, I, E, ae, A, O, U, u, 2, 3\textrhoticity, e, o}/) from the words: heed, hid, head, had, hod, hawed, hood, who'd, hud, heard, hayed, hoed.

%%%%%%%%%%%%%%%%%%%%%%%%%%%%%%%%%%%%%%%%%%%%%%%%%%%%%%%%%%%%%%%%%%%%%%%%%%%%%%

\section{METHODS}
\label{sec:methods}

Code implementing this algorithm is publicly available at \url{https://github.com/adiyoss/AutoVowelDuration}. Segments of acoustic signal for the HG dataset along with algorithmic and manual annotations, are available in the Online Speech/Corpora Archive and Analysis Resource (\url{https://oscaar.ci.northwestern.edu/}; dataset `Within Category Neighborhood Density'). For the HGCW dataset, segments of acoustic signal and manual annotations are available at: \url{http://homepages.wmich.edu/~hillenbr/voweldata.html}.

\subsection{DLM}
The direct loss minimization (DLM) algorithm described above was trained and tested on both HG and CT datasets. For the CT corpus, the training set was a set of 4049 tokens from the original corpus \citep{clopper2002indiana}, and the test set was the same 777 tokens reported in \cite{clopper2014effects}. The model parameters were tuned on a dedicated development set ($\sim 10\%$ of the training set). The parameters that yielded the lowest error rate were $\eta = 0.1$ and $\epsilon = -1.36$. We used $\taub = 1$ frame and $\taue = 2$ frames for the loss during training. The initial weight vector was set to be the averaged weight vector from the Passive-Aggressive (PA) algorithm \citep{CrammerDeKeShSi06} with $C = 0.5$ and 100 epochs. When trained and tested on the HG corpus, we used 10-fold cross validation (the reported results are the average error over all 10 folds) with the same settings and parameters as described above.

In order to better comprehend the influence of the phoneme classifier, the only language dependent feature in our system, we also trained and tested the direct loss algorithm without the acoustic features related to the phoneme classifier $G_{\text{vowel}}$, $G_{\text{nasal}}$, or $L_{\text{vowel}}$, and their corresponding feature functions. In the case when the phoneme classifier was used, we trained multiclass PA as described in \citep{DekelKeSi04a} on the TIMIT corpus of read speech.

\subsection{HMM}
To validate the effectiveness of the proposed approach, we compared it to the most common approach currently used in automatic phonetic measurement of speech: forced alignment. This is an algorithm which, given a speech utterance and its phonetic content, finds the start time of each phoneme in the speech utterance. Often the orthographic content of the speech utterance is given, and it is converted to its phonetic content using a lexicon. This procedure may not be accurate as often the surface pronunciation uttered in spontaneous speech is not the same as the canonical pronunciation that appears in the lexicon. 

Conventionally, the forced alignment is implemented by forcing the decoder of an HMM phoneme recognizer to pass through the states corresponding to the given phoneme sequence. So far, automatic vowel extraction has been done using such forced alignment procedures \citep{reddy2015toward}. While there are several open source implementations of HMMs, all the publicly available forced aligner packages \citep{goldman2011easyalign,gorman2011prosodylab,rosenfelder2014fave,yuan2008speaker} are based on the HTK toolkit \citep{Young94thehtk}. Since this is the case, we chose to compare our results with the most recent one, namely the \emph{FAVE aligner} \citep{rosenfelder2014fave},  based on the \emph{Penn Phonetics Lab Forced Aligner (P2FA)}  \citep{yuan2008speaker}. In our analyses below, we refer to this system as the HMM aligner. Note that in contrast to the DLM the forced aligner requires a phonetic or orthographic transcription as input.

Since neither CT nor HG have enough data to train an HMM forced alignment system, in this work we use a pre-trained model which was trained on different corpus (see details in \cite{rosenfelder2014fave}). In order to make a fair comparison between the proposed model and the HMM system, we evaluate the models on two different settings: (i) a dataset which was not used during training any of the models (HGCW), and (ii) on mismatched training and test datasets. More details about the experiments can be found later in the experimental results. 

%%%%%%%%%%%%%%%%%%%%%%%%%%%%%%%%%%%%%%%%%%%%%%%%%%%%%%%%%%%%%%%%%%%%%%%%%%%%%%

\section{RESULTS: MEASUREMENT DEVIATION}
\label{sec:deviation}

We first examine results on the HG and CT datasets when the DLM is algorithm is trained and tested on the same corpus. The difference between the automatic and manual measurements of vowel duration are given in Table~\ref{tab:results}, where the evaluation metric is the loss in \eqref{eq:loss} with both $\taub$ and $\taue$ equal to 0, and in Table~\ref{tab:results_prec}, where the error is given in terms of the percentage of predictions that do not fall within the boundaries of 20 ms from the manual onset and 50 ms from the manual offset. (We allow for greater deviations in vowel offset, where manual annotators often show greater disagreement.) These reveal that when trained and tested on the same corpus, the DLM generally out-performs the standard HMM approach, particularly when the phoneme classifier is incorporated into the algorithm. An exception is seen in the HG offset data, where the percentage of predictions outside of 50 ms is smallest for the HMM. 

%%%%%%%%%%%%%%%%%%%%%%%%%%%%%%%%%%%%%%%%%%%%%%%%%%%%%%%%%%%%%%%%%%%%%%%%%%%%%%

\begin{table}[h]
\centerline{
\begin{tabular}{ c c c c c c c c  }
\hline 
\hline
\multicolumn{2}{c}{} & \multicolumn{2}{c}{DLM} &  \multicolumn{2}{c}{DLM (no classifier)} & \multicolumn{2}{c}{HMM}  \\ 
\hline
\multicolumn{2}{c}{} & onset & offset & onset & offset & onset & offset  \\
\hline 
\multicolumn{2}{c}{\emph{HG}} & \textbf{5.21} & \textbf{22.80} & 5.84 & 28.98  & 16.67 & 24.72\\
\multicolumn{2}{c}{\emph{CT}} & 9.42 & \textbf{16.76} & \textbf{9.24} &  23.18 & 35.90 & 30.61\\
%\multicolumn{2}{c}{\emph{HGCW}} & 12.913 & 9.281 & 15.856 & 23.887 & 19.948 & 27.297\\
\hline           
\hline 
\end{tabular}
}
\caption{\label{tab:results} {\it Results of DLM (with and without phoneme classifier, trained and tested on the same corpus) and HMM relative to manual annotation. Average deviation of onset and offset [in msec]. Bold indicates minimum deviation for each dataset within each context.}}
\end{table}

%%%%%%%%%%%%%%%%%%%%%%%%%%%%%%%%%%%%%%%%%%%%%%%%%%%%%%%%%%%%%%%%%%%%%%%%%%%%%%
\begin{table}[h]
\vspace{2mm}
\centerline{
\begin{tabular}{c c c c c c c c c c c}
\hline
\hline 
\multicolumn{2}{c}{} & \multicolumn{2}{c}{DLM} &  \multicolumn{2}{c}{DLM (no classifier)} & \multicolumn{2}{c}{HMM} \\ 
\hline
\multicolumn{2}{c}{} & onset & offset & onset & offset & onset & offset  \\
\hline
\multicolumn{2}{c}{\emph{HG}} & \textbf{6.15\%} & 13.15\% & 8.05\% & 18.44\% & 31.90\% & \textbf{10.94\%}\\
\multicolumn{2}{c}{\emph{CT}} &\textbf{9.46\%} & \textbf{8.31\%} & 9.59\% & 9.34\% & 41.50\% & 13.14\%\\
%\multicolumn{2}{c}{\emph{HGCW}} &19.185\% & 2.158\% & 27.578\% & 6.535\% & 28.537\% & 11.691\%\\
\hline           
\hline
\end{tabular}
}
\caption{\label{tab:results_prec} {\it The percentage of predictions that do not fall within the boundaries of 20 ms at the onset and 50 ms at the offset from the manual annotation when the DLM is trained and tested on the same corpus. Bold indicates minimum deviation for each dataset within each context.}}
\end{table}

%%%%%%%%%%%%%%%%%%%%%%%%%%%%%%%%%%%%%%%%%%%%%%%%%%%%%%%%%%%%%%%%%%%%%%%%%%%%%%
It is possible that the accuracy of the algorithms may differ by consonantal context. To assess this, we classified the consonant preceding or following the vowel of each word as belonging to one of four categories: voiceless stops, voiced stops, fricatives/affricates, and sonorants. We then calculated the overall measurement error for each algorithm within each category, as reported in table \ref{tab:results_CONTEXT}.

%%%%%%%%%%%%%%%%%%%%%%%%%%%%%%%%%%%%%%%%%%%%%%%%%%%%%%%%%%%%%%%%%%%%%%%%%%%%%%
\begin{table}[h!]
\centerline{
\begin{tabular}{c c c c c c c c c c c}
\hline 
\hline
\multicolumn{2}{c}{} & \multicolumn{2}{c}{DLM} &  \multicolumn{2}{c}{DLM (no classifier)}  &  \multicolumn{2}{c}{HMM}   \\ 
\hline
\multicolumn{2}{c}{\emph{Onset}} & HG & CT & HG & CT & HG & CT\\
\hline 
\multicolumn{2}{c}{\emph{Voiceless stops}} 			& 25.67 & 17.07 & 30.65 & \textbf{16.96} & \textbf{21.95}  & 97.45	\\
\multicolumn{2}{c}{\emph{Voiced stops}}				& \textbf{27.81}& \textbf{18.62} & 31.23 & 20.27 & 31.66   & 53.90 \\
\multicolumn{2}{c}{\emph{Fricatives/affricates}} 	& 25.03 & 16.74 & 32.40 & \textbf{16.28} & \textbf{18.01} & 32.86 \\
\multicolumn{2}{c}{\emph{Sonorants}} 				& \textbf{24.12} & \textbf{26.97} & 36.32 & 27.95 & 42.35  & 40.52	\\
\hline
\multicolumn{2}{c}{} & \multicolumn{2}{c}{DLM} &  \multicolumn{2}{c}{DLM (no classifier)}  &  \multicolumn{2}{c}{HMM}   \\ 
\hline
\multicolumn{2}{c}{\emph{Coda}} & HG & CT & HG & CT & HG & CT\\
\hline 
\multicolumn{2}{c}{\emph{Voiceless stops}} 			& \textbf{25.72} & 18.06 & 28.76 & \textbf{17.78}  & 25.91 	 & 43.22 \\
\multicolumn{2}{c}{\emph{Voiced stops}}				& \textbf{16.16} & \textbf{23.08} & 20.10 & 27.22 & 19.04 & 45.19 \\
\multicolumn{2}{c}{\emph{Fricatives/affricates}} 	& \textbf{15.00} & 20.42 & 27.25 & \textbf{20.41} & 27.92 1 & 72.82 \\
\multicolumn{2}{c}{\emph{Sonorants}} 				& \textbf{32.51} & - & 39.04 & - & 35.36 & -\\
\hline           
\hline 
\end{tabular}
}
\caption{\label{tab:results_CONTEXT} {\it Results of DLM (with and without phoneme classifier, trained and tested on the same corpus) and HMM relative to manual annotation, broken down by consonant context. Onset consonant context is shown in the first, followed by coda consonant context. Average deviation of full vowel duration [in msec]. Fricatives are all voiceless, with the exception of [d\textipa{Z}] in CT coda consonants. There were no sonorants in CT codas. Bold indicates minimum deviation within each context for each dataset.}}
\end{table}

The context-specific HG data shows that the general advantage for the DLM with the phoneme classifier holds, but there are some exceptions. The performance of the HMM is comparable to the DLM with classifier when a vowel is followed by a voiceless stop, and appears to outperform the DLM when voiceless stops or fricatives precede the vowel. In the CT data, there is a clear DLM advantage across all contexts.

%%%%%%%%%%%%%%%%%%%%%%%%%%%%%%%%%%%%%%%%%%%%%%%%%%%%%%%%%%%%%%%%%%%%%%%%%%%%%%

\subsection{Mismatched training and test datasets}
The comparison above may be biased in favor of the DLM, as the algorithm's parameters may be tuned to specific features of the annotators that worked with the HG and CT corpora. To provide a more even comparison, the DLM was trained and tested on different corpora--training on HG and testing on CT and vice versa. We also present the result of our algorithm trained on CT (the manual dataset with the highest manual inter-annotator reliability; see below) and tested on the third dataset, HGCW. 

As before, the difference between the automatic and manual measurements of vowel duration are given in Table~\ref{tab:results_mismatch}, and the error in terms of the percentage of predictions that do not fall within the boundaries of 20 ms from the manual onset and 50 ms from the manual offset is given in Table~\ref{tab:results_prec_mismatch}. These results again show that for the onset data, the DLM out-performs the HMM, even when the training and testing sets for the algorithm are mismatched. In vowel offsets, the HMM meets or outperforms the DLM for the HG and CT data, but the DLM advantage is maintained for the HGCW offset data.

%%%%%%%%%%%%%%%%%%%%%%%%%%%%%%%%%%%%%%%%%%%%%%%%%%%%%%%%%%%%%%%%%%%%%%%%%%%%%%

\begin{table}[h]
\centerline{
\begin{tabular}{ c c c c c c c c c c c}
\hline 
\hline
\multicolumn{2}{c}{} & \multicolumn{2}{c}{DLM} &  \multicolumn{2}{c}{DLM (no classifier)} & \multicolumn{2}{c}{HMM}  \\ 
\hline
\multicolumn{2}{c}{} & onset & offset & onset & offset & onset & offset  \\
\hline 
\multicolumn{2}{c}{\emph{HG} (\emph{CT} model)} & \textbf{14.44} & 38.66 & 14.92 & 43.95  & 16.67 & \textbf{24.72}\\
\multicolumn{2}{c}{\emph{CT} (\emph{HG} model)} & \textbf{9.84} & 30.85 & 10.17 & \textbf{29.94} & 35.90 & 30.61\\
\multicolumn{2}{c}{\emph{HGCW} (\emph{CT} model)} & \textbf{12.91} & \textbf{9.28} & 15.86 & 23.89 & 19.95 & 27.30 \\
\hline           
\hline 
\end{tabular}
}
\caption{\label{tab:results_mismatch} {\it Results of DLM (with and without phoneme classifier) and HMM under mismatched training and test datasets. Average deviation of onset and offset [in msec].  Bold indicates minimum deviation for each dataset within each context.}}
\end{table}
%%%%%%%%%%%%%%%%%%%%%%%%%%%%%%%%%%%%%%%%%%%%%%%%%%%%%%%%%%%%%%%%%%%%%%%%%%%%%%
\begin{table}[h]
\vspace{2mm}
\centerline{
\begin{tabular}{c c c c c c c c c c c }
\hline
\hline 
\multicolumn{2}{c}{} & \multicolumn{2}{c}{DLM} &  \multicolumn{2}{c}{DLM (no classifier)}  & \multicolumn{2}{c}{HMM} \\ 
\hline
\multicolumn{2}{c}{} & onset & offset & onset & offset  & onset & offset  \\
\hline
\multicolumn{2}{c}{\emph{HG} (\emph{CT} model)} & \textbf{11.55\%} & 23.99\% & 12.25\% & 28.80\% & 31.90\% & \textbf{10.94\%} \\
\multicolumn{2}{c}{\emph{CT} (\emph{HG} model)} & \textbf{8.36\%} & 14.92\% & \textbf{8.36\%} & \textbf{10.94\%} & 41.50\% & 13.14\% \\
\multicolumn{2}{c}{\emph{HGCW} (\emph{CT} model)} & \textbf{19.18\%} & \textbf{2.16\%} & 27.58\% & 6.54\% & 28.54\% & 11.69\% \\
\hline           
\hline
\end{tabular}
}
\caption{\label{tab:results_prec_mismatch} {\it The percentage of predictions that do not fall within the boundaries of 20 ms at the onset and 50 ms at the offset from the manual annotation (with mismatched training and test datasets for the DLM). Bold indicates minimum deviation for each dataset within each context.}}
\end{table}
%%%%%%%%%%%%%%%%%%%%%%%%%%%%%%%%%%%%%%%%%%%%%%%%%%%%%%%%%%%%%%%%%%%%%%%%%%%%%%

\subsection{Measurement Correlation}
Above, we examined the degree to which each algorithm agreed with manual annotators based on the amount of deviation between measures. Another measure of  agreement conventionally reported in phonetic studies is the correlation between measurements of vowel durations. This is typically done by assigning a random subset of the data to a second annotator. For the HG dataset, the second annotator's correlation with the original annotator was r(627) = 0.84.  We performed a similar analysis, treating the algorithm as a second annotator relative to the manual labels. For the same subset of the data (with the DLM trained and tested on the same corpus), the correlations between each algorithm and the original manual annotator are DLM=0.79, DLM (no classifier)=0.64, and HMM=0.73. While using the models that were trained on  mismatched data the correlations are DLM=0.67, DLM (no classifier)=0.55. Similarly, for the CT dataset, the second annotator's correlation with the original annotator was r(398) = 0.95. For the same subset of the data, each algorithm's correlations are: DLM=0.93, DLM (no classifier)=0.93, and HMM=0.57. However, when using the models that were trained on mismatched data the correlations are DLM=0.54, DLM (no classifier)=0.53. Thus, when training the models on mismatched datasets, correlations of the HMM model and DLM model with human annotators are equivalent; the DLM only outperforms HMM when using the same training and testing set.

%%%%%%%%%%%%%%%%%%%%%%%%%%%%%%%%%%%%%%%%%%%%%%%%%%%%%%%%%%%%%%%%%%%%%%%%%%%%%%

\subsection {Discussion}
Analysis of measurement deviation suggests our model generally outperforms the HMM-forced alignment algorithm, even when training and testing sets are mismatched. Incorporating the phoneme classifier improved performance, but even without the classifier the DLM typically outperforms the HMM.  However, with respect to measurement correlation, there was no clear advantage; DLM and HMM exhibited equivalent performance when trained on mismatched training and test sets.

\section{RESULTS: REPRODUCING REGRESSION MODEL FINDINGS}
\label{sec:evaluation_regression}

Empirical studies of speech and language processing use acoustic properties such as vowel duration as behavioral measures of the effects of various types of variables that influence speech and language. Canonical examples of variables of study include the phonetic context of vowels (e.g. preceding voiced vs. voiceless stops), properties of speakers who produce those vowels (e.g. native vs. non-native speakers) and the linguistic (lexical, syntactic, etc.) context in which the vowels appear (e.g. predictable vs. unpredictable). The effects of these variables on acoustic properties are typically examined using inferential statistical models. 

The second evaluation metric of our algorithms therefore examined the similarity of inferential statistical model fits based on measurements generated by algorithms vs. manual measurements. Out of the three datasets, only HG constructed models based on duration data; our analysis therefore focused on this dataset. (In supplemental materials, we examined the CT dataset, providing the vowel duration measurements of our algorithm as well as the HMM as inputs to a formant estimator.)

\cite{heller2014grammatical} examined whether processes involved in sentence planning influence the processing of sound. Speakers named pictures in a context that strongly emphasized sentence planning (following a sentence fragment) as well as a context that did not require substantial planning (producing the picture name in isolation). The order of these contexts was counterbalanced across speakers. To index effects of sound structure processing, this study also manipulated the number of words phonologically similar to the target that share its grammatical category (category-specific lexical density). 

To analyze the effect of this variables, HG used linear mixed effects regression models \citep{baayen2008mixed}, an approach that has become dominant in the analysis of speech and psycholinguistic data. These regression models predict dependent measures based on a linear combination of predictors, including both fixed and randomly distributed predictors capturing variation in effects by both participants and items. This allows researchers to examine the effects of interest while controlling for other properties of the words and speakers.

Analysis of the full data set showed that lexical density and vowel category had no effect on vowel durations \citep{heller2014grammatical, heller2015grammaticalCorrigendum}; however, these effects were not stable when the subset of data used here was examined. Therefore, the models used here were simplified from those reported in the original paper, including two contrast-coded fixed effect factors: production context (isolation vs. sentence) and block (first vs. second). To control for contributions from the random sample of participants and items used in this experiment (compared to all English speakers and all words of English), we included two sets of random effects. Random intercepts for both speakers and words were included, along with uncorrelated slopes for context by both speaker and word.

\subsection{Significance of fixed effects in models of the HG dataset}

We first examine the primary interest of many phonetic studies--the binary distinction between significant vs. insignificant effects of fixed-effect predictors (e.g., the effect of experimental condition). 
To control for skew, vowel durations were log-transformed prior to analysis. Given that outliers can have an outsized influence on parameter estimates, all regressions models were refit after excluding observations with standardized residuals exceeding 2.5 \citep{baayen2008analyzing}. 
The significance of fixed-effects predictors was assessed by using the likelihood ratio test to compare models with and without the predictor of interest \citep{barr2013random}. Table \ref{tab:hellerFixedEffects} compares the estimates for the two fixed effects parameters for the manual model and each algorithmic model. Although there was some variation in the model estimates of the fixed effects for each algorithm compared to the manual annotations, each algorithm recovered the overall pattern of a significant effect of context but not block.

\begin{table}
	\centering
	\def\arraystretch{1.25}
	\begin{tabular}{@{\extracolsep{\fill} }l c c c c}
		\hline\hline & Context & \textit{t} & Block & \textit{t}\\
		\hline 
		Manual & \textbf{-0.057 (0.017)} & \textbf{-3.29} &  0.012 (0.016) & 0.72 \\
		DLM & \textbf{-0.072 (0.019)} & \textbf{-3.82} & 0.016 (0.018)  & 0.91 \\
		DLM (no classifier) & \textbf{-0.059 (0.018)} & \textbf{-3.34} & 0.015 (0.017) & 0.90 \\
		HMM & \textbf{-0.058 (0.018)} & \textbf{-3.13} &  0.008 (0.014) &  0.55 \\
		\hline  \hline
	\end{tabular}
	\caption[Regression model fixed effects, Heller \& Goldrick dataset]{Estimates and \textit{t}-values for fixed effects in regression model, Heller \& Goldrick dataset (standard error of estimate in parentheses). Significant effects (\textit{p} $<=$ 0.05, as assessed by likelihood ratio tests) are bolded. }
	\label{tab:hellerFixedEffects}
\end{table}

\subsubsection{Comparison of predictions of models of the HG dataset} 
\begin{figure}[h]
	\centering
	\includegraphics[width=0.6\linewidth]{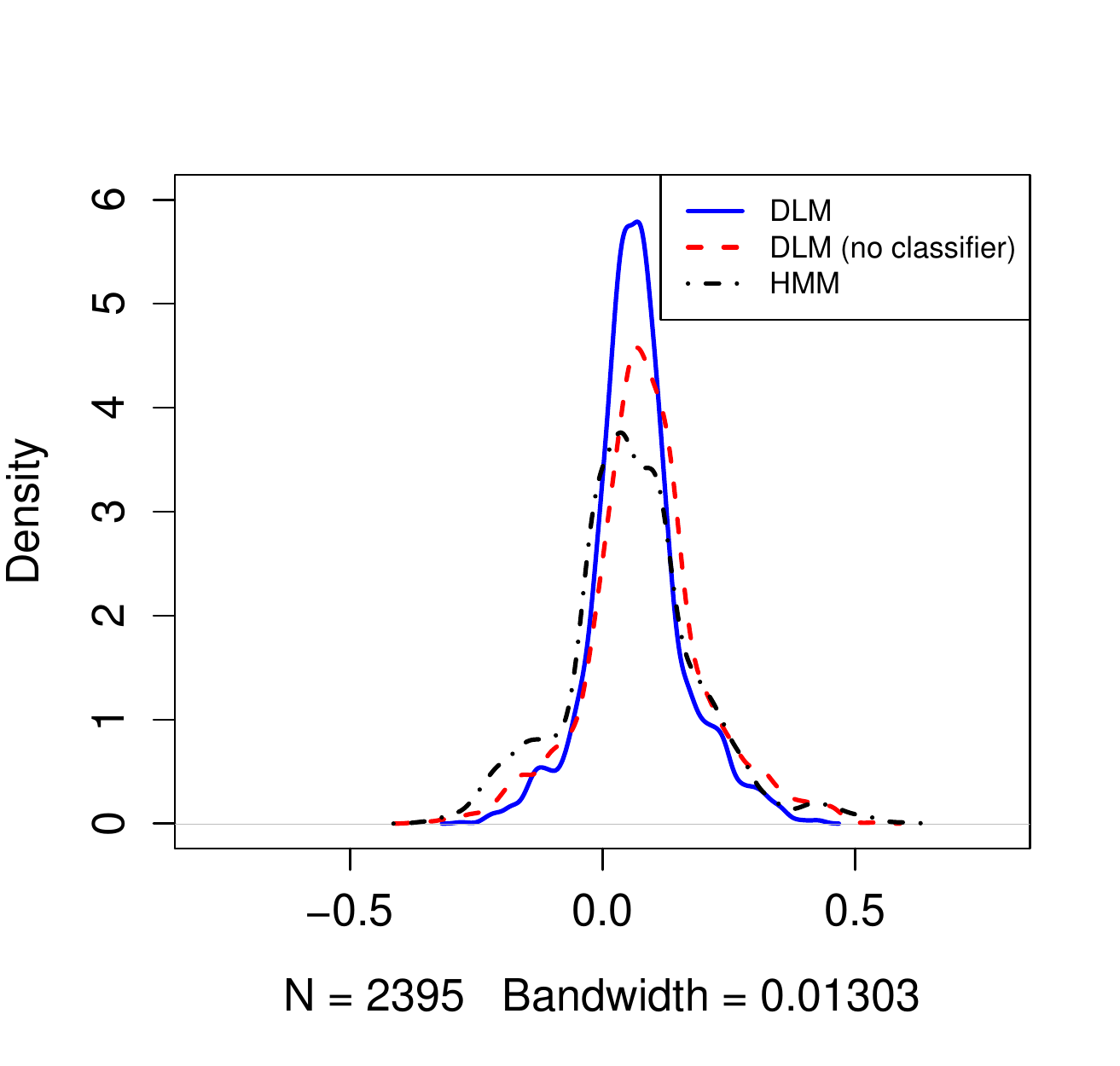}
	\caption{Comparisons of leave-one-out model predictions of vowel durations for the Heller \& Goldrick (2014) dataset. Deviance of each algorithmic method as compared to the manual predictions (manual-algorithmic) are plotted as individual lines.}
	\label{fig:hellerLOODensity}
\end{figure}

An alternative, more global, assessment of model similarity is to compare model predictions. This was assessed by leave-one-out validation. For each observation, we excluded it from the dataset and re-fit the regression to the remaining observations. After residual-based outlier trimming (and re-fitting), we examined the predictions of this re-fitted model for the excluded observation. Figure \ref{fig:hellerLOODensity} shows distributions of the deviation of each algorithm's predicted fit to the predicted model fit to the manual data.

%%%%%%%%%%%%%%%%%%%%%%%%%%%%%%%%%%%%%%%%%%%%
%%%%%%%%%%%%%%%%%%%%%%%%%%%%%%%%%%%%%%Figure 4
%%%%%%%%%%%%%%%%%%%%%%%%%%%%%%%%%%%%%%%%%%%%
%%%%%%%%%%%%%%%%%%%%%%%%%%%%%%%%%%%%%%%%%%%%

The DLM algorithm out-performed the other algorithms. The mean squared error relative to the predictions of the model fit to the manually annotated data was lowest for the DLM algorithm (0.479 msec$^{2}$), higher for HMM (0.835 msec$^{2}$), and larger still for DLM (no classifier) (0.935 msec$^{2}$). Bootstrap confidence intervals of the differences across algorithms showed that DLM outperformed the other two algorithms (95\% CI of differences from HMM: [0.00039, 0.00043] msec$^{2}$; DLM (no classifier): [0.00037, 0.00055] msec$^{2}$). The HMM algorithm did not significantly differ from DLM (no classifier) (95\% CI [-0.000200, 0.000002] msec$^{2}$).

Parallel to the analysis of measurement deviation, these analyses suggest our model outperforms the HMM-forced aligner. Incorporating the phoneme classifier improved performance, but even without the classifier the performance of the DLM meets or exceeds that of the HMM.

%%%%%%%%%%%%%%%%%%%%%%%%%%%%%%%%%%%%%%%%%%%%%%%%%%%%%%%%%%%%%%%%%%%%%%%%%%%%%%

\section{GENERAL DISCUSSION}\label{sec:discussion}
 
We presented an algorithm for automatically estimating the durations of vowels based on the structured prediction framework, relying on a set of acoustic features and feature functions finely tuned to reflect the properties of vowel acoustics relevant to vowel segmentation. The DLM algorithm, when including a phoneme classifier, clearly succeeds at matching manual measurements of vowel duration. With respect to both measurement deviation and reproduction of regression model results, the DLM algorithm out-performs or matches a commonly-used forced alignment system while not requiring a phonetic transcription. This approach can allow laboratory experiments to address much larger samples of data in a way that is replicable and reliable.

Having achieved some success with monosyllabic, laboratory stimuli, future development of this algorithm should extend this approach to more naturalistic production. In current work, we are extending this approach to vowel durations in multisyllabic words. We believe that moving outside the laboratory will be facilitated by the structure of our approach; in contrast with existing systems, our algorithm has the additional benefit of not requiring a transcript of the desired speech prior to analysis. Extending the capability of the system to analyze untranscribed speech will support the analysis of more naturalistic, connected speech, including speech styles or dialects that may be difficult to robustly sample in a laboratory context \citep{labov1972sociolinguistic, rischel1992formal}. 

With respect to the algorithms utilized in future work, we aim to apply new advances in deep learning to automatic estimation of vowel duration. In preliminary work, we used this approach to train a binary classifier to detect for each frame whether it contains a vowel or not \citep{AdiKeGo15}. However, since this approach does not take into account the relationships between adjacent frames, it requires an additional algorithm to yield durations. In future work, we plan to focus on combining new advances in sequence deep learning \citep{elman1990finding,graves2013speech,graves2014towards} with our current structured prediction scheme.

\bigskip

\noindent \textbf{Acknowledgements}
 
\setlength{\parindent}{0.7cm} 

Research supported by NIH grant 1R21HD077140 and NSF grant BCS1056409.

%\clearpage
%\noindent
%\textbf{Figure Captions}\\
%\noindent
%Figure 1. An example of our notation. The top panel presents the signal in the time domain, and the middle panel presents the spectrogram of the signal. The vertical solid lines present the annotated vowel's onset $\tb$ and offset $\te$. The speech signal is represented by a sequence of acoustic feature vectors as depicted in the lower panel. For the $t$-th frame, the acoustic feature vector is denoted by $\vx_t$.\\
%\noindent
%Figure 2. Direct loss minimization training procedure.\\
%\noindent
%Figure 3. Values of acoustic features for an example acoustic sample (the word ``got''). The vertical dashed lines indicate the annotated onset and offset of the vowel (color online).\\
%\noindent
%Figure 4. Comparisons of leave-one-out model predictions of vowel durations for the Heller \& Goldrick (2014) dataset. Deviance of each algorithmic method as compared to the manual predictions (manual-algorithmic) are plotted as individual lines.\\

\end{document}

%%% TITLE PAGE
\begin{titlepage}
\begin{center}

\textbf{Supplemental Materials:}\\
\textbf{Automatic measurement of vowel duration via structured prediction}\\

\vspace{5ex}

Yossi Adi\footnote{e-mail: adiyoss@cs.biu.ac.il}, 
Joseph Keshet\footnote{e-mail: joseph.keshet@biu.ac.il} \\
Department of Computer Science, Bar-Ilan University, Ramat-Gan, Israel,  52900\\
\vspace{5ex}
Emily Cibelli\footnote{e-mail: emily.cibelli@northwestern.edu},
Erin Gustafson\footnote{e-mail:egustafson@u.northwestern.edu}\\
Department of Linguistics, Northwestern University, Evanston, IL, 60208\\
\vspace{5ex}
Cynthia Clopper\footnote{e-mail:clopper.1@osu.edu}\\
Department of Linguistics, Ohio State University, Columbus, OH, 43210\\
\vspace{5ex}
Matthew Goldrick\footnote{e-mail: matt-goldrick@northwestern.edu}\\
Department of Linguistics, Northwestern University, Evanston, IL, 60208\\
\end{center}
\markboth{\hfill Automatic vowel measurement:Supplemental data \ }{\hfill Automatic vowel measurement:Supplemental data \ }

\end{titlepage}

\addtocounter{page}{1}

%%%%%%%%%%%%%%%%%%%%%%%%%%%%%%%%%%%%%%%%%%%%%%%%%%%%%%%%%%%%%%%%%%%%%%%%%%%%%%

\section{RESULTS: REPRODUCING REGRESSION MODEL FINDINGS USING A FORMANT ESTIMATOR}
\label{sec:evaluation_regression}

In the main text, we report analyses of vowel duration, which the DLM and HMM algorithms are designed to directly estimate. In these materials, we examine the similarity of inferential statistical model fits based on measurements generated by a standard formant estimator using as input vowel durations from algorithms vs. manual measurements. 

\cite{clopper2014effects} examined how degree of vowel contrast was influenced by two factors: lexical contrast (the presence vs. absence of a word in English with the other member of the vowel); and dialect (Northern speakers have a smaller contrast than Midland speakers for /\textipa{E}\mytilde   \textipa{ae}/, whereas the pattern is reversed for  /\textipa{A}\mytilde \textipa{O}/).

Vowel contrast was quantified by the distance in Bark \citep{traunmuller1990analytical} between the first and second formants (F1/F2) of vowels from contrasting categories. These were estimated at vowel midpoint using Praat \citep{boersma2001praat}. Vowel distance was defined as the median of the Euclidean distances in the F1-F2 Bark space from a given vowel to each corresponding vowel in the same competitor condition. For example, if the target word was \textit{dead}, the relevant set of distances would be to /\textipa{ae}/ target words with lexical competitors (e.g., \textit{dad}). The effect of lexical competitor and dialect on these distance values was modeled via treatment-coded fixed effects in a mixed-effects regression model by \cite{clopper2014effects,clopper2015erratum}. Lexical neighborhood density and log word frequency were included as control factors. Random intercepts for word and talker were included, along with uncorrelated random slopes for density and frequency by talker. 

Our analysis followed that of the original paper, except that all fixed-effect factors were centered (resulting in contrast-coded categorical factors), the exclusion of outliers with large model residuals, and the assessment of significance via model comparison (see below). Using each algorithm's segmentation of the vowel, Praat was used to calculate F1 and F2 values. Tokens with F1 or F2 values more than 3 standard deviations from the mean of each vowel were considered to be outliers. As the authors in the original paper replaced outlier tokens with manual measurements, we employed a similar approach; for each token identified as an outlier (5.7\% of tokens), the distance measurements were replaced with the manual measurements corresponding to those tokens. 

\section{Significance of fixed effects}

Following the HG analyses, models were refit after excluding observations with standardized residuals exceeding 2.5 \citep{baayen2008analyzing} and the significance of fixed-effects predictors was assessed via the likelihood ratio test \citep{barr2013random}. 

The parameter estimates for model fit to the subset of the vowel distance data contrasting /\textipa{ae}\mytilde\textipa{E}/ are shown in Table \ref{tab:clopperFixedEffectsAE}. The manual model had a significant effect of dialect ($\chi^{2}$ = 5.05, \textit{p} = 0.025) but no effect of lexical competitor ($\chi^{2}$ = 0.11, \textit{p} = 0.074). The interaction of dialect and competitor was not significant ($\chi^{2}$ = 0.09, \textit{p} = 0.7631). 

The algorithms largely recovered the same effects as found in the manual data. 
The DLM model recovered the significant main effect of dialect ($\chi^{2}$ = 5.04, \textit{p} = 0.025); this parameter was marginal in the DLM (no classifier) model ($\chi^{2}$ = 3.09, \textit{p} = 0.079) and the HMM model ($\chi^{2}$ = 3.66, \textit{p} = 0.056). The interaction did not reach significance for any of the algorithmic models, but was marginal in the DLM model ($\chi^{2}$ = 2,83, \textit{p} = 0.093). As in the manual model, all algorithmic models found no significant or marginal effects for the lexical competitor factor or either control variable (\textit{p} $>$ 0.10).

\begin{sidewaystable}\scriptsize
%\begin{table}[t]
	\centering
	\begin{tabular}{@{\extracolsep{\fill} }l c c c c c c c c c c}
		\hline\hline
		 & Frequency & \textit{t} & Density & \textit{t} & Dialect & \textit{t} & \thead{Lexical\\Competitor}& \textit{t} & \thead{ Dialect *\\ Competitor} & \textit{t} \\
		% frequency, density, dialect, condition, interaction
		\hline 
		Manual & 0.026 (0.085) & 0.302 & -0.001 (0.007) & -0.115 & \textbf{-0.345  (0.144)} & \textbf{-2.390} &  0.039 (0.114)  & 0.338 & 0.114 (0.078) & 1.462 \\
		
		DLM &  0.0357 (0.119) &  0.301 & 0.001 (0.011) &  0.054
		 & \textbf{-0.355 (0.148)} & \textbf{-2.396} & 0.091 (0.175) & 0.519 & \textit{0.138  (0.0818)} & \textit{1.687}\\
		 
		\makecell{DLM (no \\ classifier)} & 0.014 (0.123) &  0.113 & 0.001 (0.011) &  0.112 & \textit{-0.277 (0.151)} &  \textit{-1.828} & 0.111 (0.182)  & 0.60 & 0.131 (0.083)  & 1.586 \\
		
		HMM & \textit{0.1345 (0.102)} & \textit{ 1.322} & -0.003 (0.009) & -0.298 & \textit{-0.257 (0.128)} & \textit{-2.003} & -0.121   (0.144) &  -0.839 & -0.007 (0.089)  & -0.074\\
		
		\hline\hline
	\end{tabular}
	\caption[Regression model fixed effects, Clopper \& Tamati dataset, /\textipa{ae}\mytilde\textipa{E}/data.]{Estimates and \textit{t}-values for fixed effects in regression model, Clopper \& Tamati dataset (standard error of estimate in parentheses), /\textipa{ae}\mytilde\textipa{E}/ data. Significant effects (\textit{p} $<=$ 0.05, as assessed by likelihood ratio tests) are bolded; marginal effects (0.05 $>$ \textit{p} $<$ 0.10) are italicized.}
	\label{tab:clopperFixedEffectsAE}
%\end{table}
\end{sidewaystable}

The parameter estimates for model fit to the subset of the data contrasting /\textipa{A}\mytilde\textipa{O}/ are shown in Table \ref{tab:clopperFixedEffectsC}. The manual model had a significant effect of lexical competitor ($\chi^{2}$(1) = 12.40, \textit{p} $<$ 0.0001), but the effect of dialect was not significant ($\chi^{2}$(1) = 2.48, \textit{p} = 0.115). The interaction of dialect and competitor just missed reaching significance ($\chi^{2}$(1) = 3.65, \textit{p} = 0.056).

Models from all algorithmic methods produced a similar, but not identical, pattern of fixed effects as those found in the manual model. The effect of lexical competitor was significant for all methods (all $\chi^{2}$(1) $>$ 7, \textit{p} $<$ 0.05). The effect of dialect did not reach significance in the HMM model ($\chi^{2}$(1) = 1.14, \textit{p} = 0.285) but was significant in the DLM model ($\chi^{2}$(1) = 5.07, \textit{p} = 0.024) and marginal in the DLM (no classifier) model ($\chi^{2}$(1) = 3.54, \textit{p} = 0.060). While the interaction of dialect and competitor just missed reaching significance in the manual model it was significant in the models of all three algorithmic methods (all $\chi^{2}$(1) $>$ 3.9, \textit{p} $<$ 0.04). 

\begin{sidewaystable}\scriptsize
	\centering
	\begin{tabular}{@{\extracolsep{\fill} }l c c c c c c c c c c}
		\hline\hline
		 & Frequency & \textit{t} & Density & \textit{t} & Dialect & \textit{t} & \thead{\scriptsize Lexical\\ \scriptsize Competitor}& \textit{t} & \thead{\scriptsize Dialect *\\ \scriptsize Competitor} & \textit{t} \\
		% frequency, density, dialect, condition, interaction
		\hline 
		
		Manual & -0.056 (0.071) & -0.791 & -0.009 (0.009) & -0.943 & -0.218 (0.134) & -1.627 & \textbf{0.555 (0.132)} &  \textbf{4.194} & \textit{0.160   (0.083)}  & \textit{1.925} \\
		
		DLM & -0.059 (0.070) & -0.841  & -0.007  (0.010) & -0.714  &  \textbf{-0.256 (0.107)} & \textbf{-2.404} & \textbf{0.385 (0.129)} &  \textbf{2.977} &  \textbf{0.190  (0.089)}  & \textbf{2.147} \\
		
		\makecell{DLM (no \\ classifier)} & -0.065 (0.067) & -0.963 & -0.009 (0.009) & -0.941 & \textit{-0.220 (0.112))} & \textit{-1.969} & \textbf{0.481   (0.125)} & \textbf{3.842} &  \textbf{0.180 (0.090)}  & \textbf{2.003}  \\
		
		HMM & -0.025 (0.064) & -0.387 & -0.007 (0.008) & -0.807 & -0.156 (0.144) & -1.083 & \textbf{ 0.377 (0.117)} & \textbf{3.223} & \textbf{0.215 (0.096)} & \textbf{2.249}  \\
		
		\hline \hline
	\end{tabular}
	\caption[Regression model fixed effects, Clopper \& Tamati dataset, /\textipa{A}\mytilde\textipa{O}/ data.]{Estimates and \textit{t}-values for fixed effects in regression model, Clopper \& Tamati dataset (standard error of estimate in parentheses), /\textipa{A}\mytilde\textipa{O}/ data. Significant effects (\textit{p} $<=$ 0.05, as assessed by likelihood ratio tests) are bolded; marginal effects (0.05 $>$ \textit{p} $<$ 0.10) are italicized.}
	\label{tab:clopperFixedEffectsC}
\end{sidewaystable}

\section{Comparison of model predictions}
\label{sec:clopperLOO}

\begin{figure*}	
	\centering
	\includegraphics[width=0.99\linewidth]{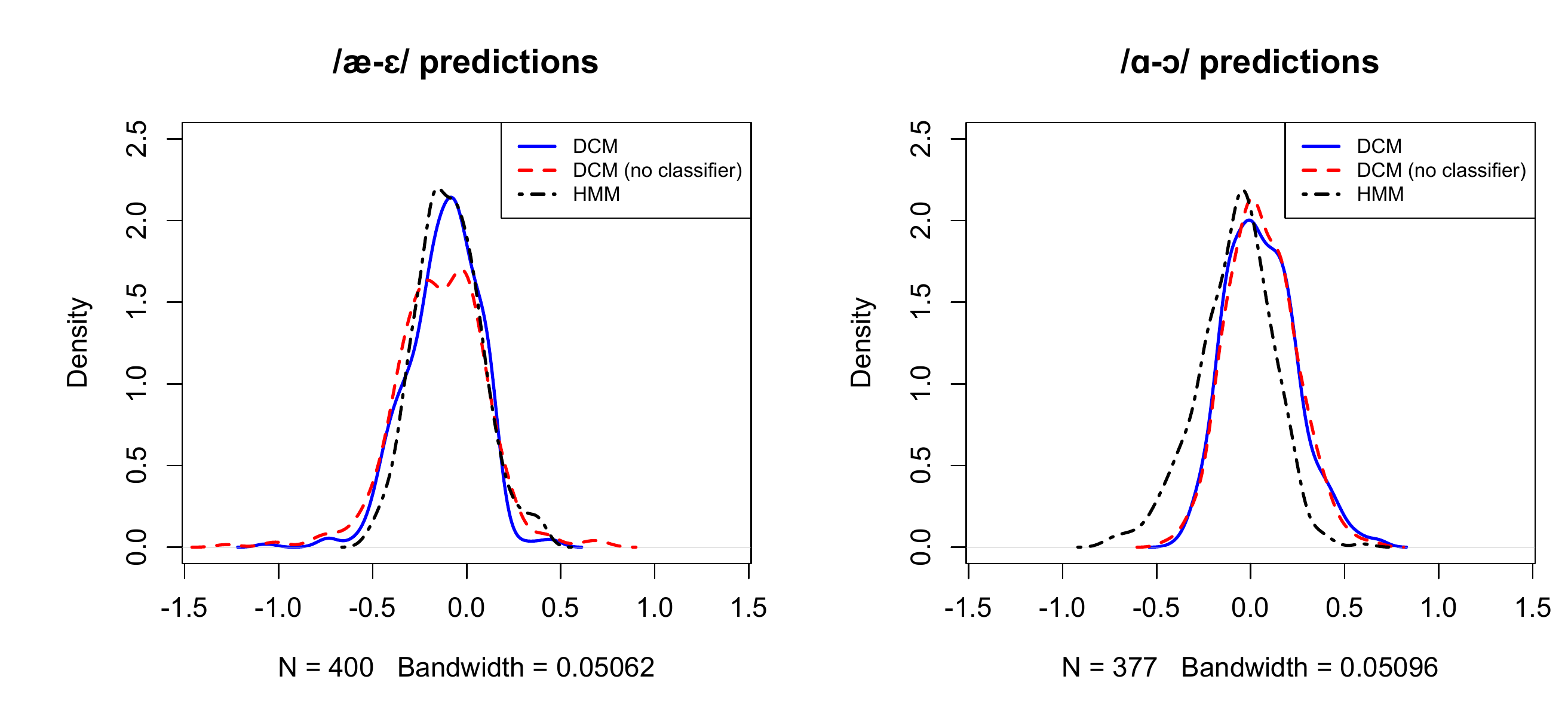}
	\caption[Predictions from leave-one-out models, Clopper \& Tamati distance data by vowel pair]{Comparisons of leave-one-out model predictions of vowel distances for the Clopper \& Tamati (2014) dataset. In each panel, the deviance of each algorithmic method as compared to the manual predictions (manual-algorithmic) are shown. Predictions for the /\textipa{ae}\mytilde\textipa{E}/ distances are presented on the left, and /\textipa{A}\mytilde\textipa{O}/ distances on the right.}
	\label{fig:clopperLOODensity}
\end{figure*}

We performed leave-one-out validation on the /\textipa{ae}\mytilde\textipa{E}/ and /\textipa{A}\mytilde\textipa{O}/ models. Comparisons of the predictions of each algorithmic model fit to the manual data predictions are shown in Figure \ref{fig:clopperLOODensity}.

For the /\textipa{ae}\mytilde\textipa{E}/ data predictions, the mean squared error relative to the manual data predictions was lowest for the HMM model (0.045 Bark$^{2}$), followed by the DLM model (0.047 Bark$^{2}$) and the DLM model without classifier (0.069 Bark$^{2}$). Bootstrap confidence intervals of the differences between each system's predictions found that the HMM system had a smaller mean squared error than the DLM system without classifier (95\% CI, [-0.033, -0.015] Bark$^{2}$); the DLM predictions also had a smaller mean squared error than the DLM without classifier predictions (95\% CI, [-0.027, -0.017] Bark$^{2}$). There was no difference between the HMM and DLM predictions (95\% CI, [-0.010, 0.005] Bark$^{2}$).

Turning to the /\textipa{A}\mytilde\textipa{O}/ data predictions, the mean squared error relative to the predictions of the manually annotated data was similar across all methods. The bootstrap confidence intervals of the differences between each algorithm's predicted values found no differences between the predictions of any of the algorithmic methods; all confidence intervals contained 0 (DLM vs. HMM 95\% CI: [-0.019, 0.004] Bark $^{2}$; DLM no classifier vs. HMM 95\% CI: [-0.018, 0.008] Bark $^{2}$; DLM vs. DLM no classifier 95\% CI: [-0.002, 0.006] Bark $^{2}$). 

\section{DISCUSSION}\label{sec:discussion}

The formant estimator used here returns relatively similar results for input based on manual vs. algorithmic measurements. However, there is some degree of variance across methods, which is somewhat unexpected given that vowel measurements are targeted at the steady-state midpoint of each vowel. Use of more robust and reliable formant estimators might provide a better comparison across different analysis methods.

\noindent \textbf{Acknowledgements}
 
\setlength{\parindent}{0.7cm} 

Research supported by NIH grant 1R21HD077140 and NSF grant BCS1056409.

\bibliographystyle{abbrvnat}
\bibliography{supplemental}